# Boolean network robotics: a proof of concept

Andrea Roli*and Mattia Manfroni
DEIS-Cesena
*Alma Mater Studiorum* Università di Bologna, Italy

Carlo Pinciroli and Mauro Birattari
IRIDIA
Université libre de Bruxelles, Belgium

June 4, 2018

## Abstract

Dynamical systems theory and complexity science provide powerful tools for analysing artificial agents and robots. Furthermore, they have been recently proposed also as a source of design principles and guidelines. Boolean networks are a prominent example of complex dynamical systems and they have been shown to effectively capture important phenomena in gene regulation. From an engineering perspective, these models are very compelling, because they can exhibit rich and complex behaviours, in spite of the compactness of their description. In this paper, we propose the use of Boolean networks for controlling robots' behaviour. The network is designed by means of an automatic procedure based on stochastic local search techniques. We show that this approach makes it possible to design a network which enables the robot to accomplish a task that requires the capability of navigating the space using a light stimulus, as well as the formation and use of an internal memory.

## 1 Introduction

Dynamical systems provide metaphors and tools which can be effectively used for analysing artificial agents,[1] such as robots. For example, gait patterns in robots can be identified and classified by analysing robots as dynamical systems and associating each gait pattern with a system attractor [10]. The dynamical systems metaphor has also been advocated as a powerful source of design principles for robotics [19]. The core idea supporting this viewpoint is that information processing can be seen as the evolution in time of a dynamical system [25]. In this paper, we propose the use of Boolean networks, a genetic regulatory network model, as robotic programs.[2] Genetic regulatory networks (GRNs) model the interaction and dynamics among genes. From an engineering and computer science perspective, GRNs are extremely interesting because they are capable of producing complex behaviours, notwithstanding the compactness of their description. For this reason, we believe that GRNs can effectively play the role of robot programs. The design of GRNs characterised by a given dynamics is a hard task *per se* and utilising such systems as robot programs is even harder, in that the GRN is required to interact with the environment, which possesses its own dynamics. The complexity of this task, though, can be tamed through the use of automatic design procedures, such as optimisation algorithms. Automatic design procedures can make the process more robust and general with respect to a customised procedure. In addition, automatic design procedures have been proven to be effective in exploring huge design search spaces and in finding innovative design solutions.[3] In this work, a Boolean network is designed so as to serve as a program for a real robot whose goal is to alternate *phototaxis* and *antiphototaxis* behaviours (i.e., going toward the positive or negative gradient of a light stimulus, respectively) depending on an external sound signal. Despite its simplicity, the robot task is not trivial, because it requires the robot to find and seek or escape from the light, and to have memory of the action being performed. We selected this task because it is a typical benchmark in evolutionary robotics [6].

---

*Corresponding author. Email: `andrea.roli@unibo.it`

[1] We use the term *agent* in its broadest general meaning.

[2] Following Russell & Norvig [23], we call the *robot program* the computational model of the system that maps the sensor readings of the robot—or agent—to the actions it takes, possibly according to an utility function and a goal.

The remainder of this paper is structured as follows. In Section 2, Boolean networks are introduced, and the methodology used for designing such systems is illustrated in Section 3. The test case is described and discussed in Section 4. Conclusions and an outlook to future work are given in Section 5.

## 2 Boolean networks

Boolean networks (BNs) have been introduced by Kauffman [11, 14] as a GRN model. BNs have been proven to reproduce very important phenomena in genetics and they have also received considerable attention in the research communities on complex systems [2, 14]. A BN is a discrete-state and discrete-time dynamical system whose structure is defined by a directed graph of $N$ nodes, each associated to a Boolean variable $x_i$, $i = 1, \ldots, N$, and a Boolean function $f_i(x_{i_1}, \ldots, x_{i_{K_i}})$, where $K_i$ is the number of inputs of node $i$. The arguments of the Boolean function $f_i$ are the values of the nodes whose outgoing arcs are connected to node $i$ (see Figure 1a). The state of the system at time $t$, $t \in \mathbb{N}$, is defined by the array of the $N$ Boolean variable values at time $t$: $s(t) \equiv (x_1(t), \ldots, x_N(t))$. The most studied BN models are characterised by having a *synchronous* dynamics—i.e., nodes update their states at the same instant—and *deterministic* functions (see Figure 1b). However, many variants exist, including asynchronous and probabilistic update rules [26].

BN models' dynamics can be studied by means of usual dynamical systems methods [4, 25], hence the usage of concepts such as state (or phase) space, trajectories, attractors and basins of attraction. BNs can exhibit complex dynamics and some special ensembles have been deeply investigated, such as that of Random BNs. Recent advances in this research field, along with efficient mathematical and experimental methods and tools for analysing BN dynamics, can be mainly found in works addressing issues in GRNs or investigating properties of BN models [1, 9, 20, 24]. These methods make it possible to analyse network dynamics and thus have insight into the behaviour of a BN system.

---

[3]A notable example is that of *evolutionary robotics* [17].

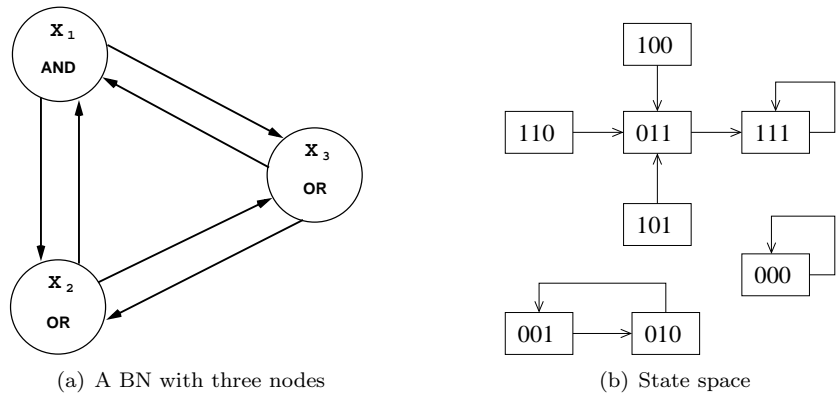

(a) A BN with three nodes    (b) State space

Figure 1: An example of a BN with three nodes (a) and its corresponding state space under synchronous and deterministic update (b). The network has three attractors: two fixed points, $(0,0,0)$ and $(1,1,1)$, and a cycle of period 2, $\{(0,0,1),(0,1,0)\}$.

## 3   Boolean network robotics

In this paper, we advocate and promote the use of BNs—and GRNs in general—to design robotic and multi-agent systems. We believe that the strengths of this approach stem from the richness of possible dynamics, along with adaptiveness and robustness, which BNs exhibit. These properties can be exploited in the context of robotic systems engineering. Moreover, theoretical and experimental tools are available for both analysing and designing these systems. We would like to emphasise that we do not claim the superiority of this approach over any other, but rather we aim at enriching methods and methodologies for designing and analysing robotic systems and intelligent systems in general. The approach we propose consists in using one or more BNs as robot program. In this way, robot dynamics can be described in terms of trajectories in a state space, making it possible to design the robot program by directly exploiting the dynamical characteristics of BNs, such as their attractors, basins of attraction and any dynamical property in general.

The design of a BN robotic system involves several interrelated tasks, which depend upon the design goal and which can be combined in several ways. The discussion of the issues arising in the design of such systems is beyond the scope of this brief communication and here we just outline the main ones.

The first task concerns the coupling of the BN with the robot, i.e., the definition of the mapping between sensors and network's inputs and between network's outputs and actuators. BNs are usually considered as isolated systems, as they are not assumed to have inputs. However, some notable exceptions exist [3, 7, 13, 18]. We assume that the values of a set of nodes (BN input nodes) are imposed by sensor readings and the values of another set of nodes (BN output nodes) are observed and used to encode the signals for maneuvering the robot's actuators. Figure 2 shows the scheme of the coupling between BN and robot. The most natural way of defining this mapping is via a direct encoding, but more elaborated ways are possible; for example, the mapping could also be defined by a learning process.

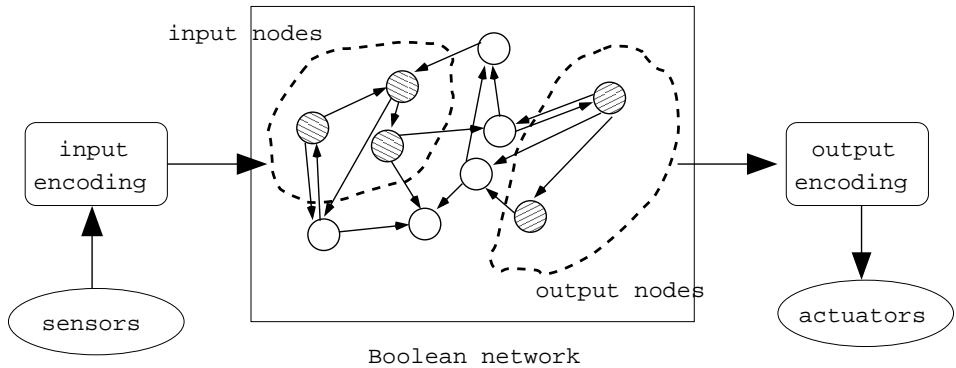

Figure 2: The coupling between BN and robot.

Once the input and the output mappings are defined, the BN that forms the robot program has to be designed. A possible way for achieving this goal is to design a BN such that its dynamics satisfies given requirements. For example, we may want to design a BN such that its attractors with largest basins of attraction correspond to the high-level behaviours the robot must exhibit and the transitions between attractors would correspond to transitions between behaviours. In this way, the dynamics of the network is directly mapped into the behaviour of the robot. Another possibility consists in modelling the BN design process as a *search problem*, in which the goal is maximising the robot's performance. These two ways are not alternative and can be combined. For example, once the basic behaviour is obtained with the latter approach, the former one is followed in order to improve the robot's behaviour—e.g., by enlarging the basin of attraction of a relevant attractor or improving network robustness.

## 3.1 Design methodology

We propose the use of a design methodology based on metaheuristics. In fact, the design of a BN that satisfies given criteria can be modeled as a constrained combinatorial optimisation problem by properly defining the set of decision variables, constraints and the objective function. This approach is illustrated by the scheme in Figure 3. The metaheuristic algorithm manipulates the decision variables which encode structure and Boolean functions of a BN. A complete assignment to those variables defines an instance of a BN. This network is then simulated and evaluated according to the specific target requirements, either on its dynamics or on the robot's behaviour, or both. A specific software component is devoted to evaluate the BN and returns an objective function value to the metaheuristic algorithm, that, in turn, proceeds with the search.

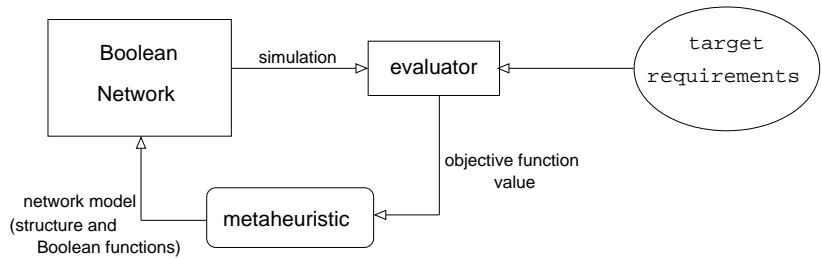

Figure 3: BN design by metaheuristics.

Despite the number of analytical studies on the properties of BNs, little effort has been devoted to their synthesis. The first study on the automatic design of BNs has been presented by Kauffman [12]. In that work, an evolutionary algorithm is applied to evolve BNs with an attractor containing a target state. A follow-up of that seminal work is that of Lemke at al. [15]. More recently, works addressing the evolvability of robustness in BNs have been presented [1, 5, 27]. In the same direction is a recent paper, in which the global fitness function is defined as the sum of single functions, each related to a network parameter linked to network robustness [8]. Finally, Roli et al. present a study on the impact of the characteristics of the BNs composing the initial population over the performance of a genetic algorithm whose goal is to design networks with a given attractor length [21].

## 4 Test case: phototaxis and antiphototaxis

The case study presented in this work consists of a robot that selects actions with respect to a light source and a sound signal. The robot must be able to perform two different behaviours: going towards the light (*phototaxis*) or moving away from it (*antiphototaxis*). In beginning of the experiment, the robot must perform phototaxis; subsequently, it must switch its behaviour to antiphototaxis after perceiving a sharp sound (like hand clapping) which is triggered at a random instant during the experiment. Therefore, the robot needs somehow to keep memory of the perception of the clap to select the action to be performed at any instant in time. The robot is first designed and tested in simulation; afterwards, its performance is assessed in a real physical setting by using an *e-puck* robot [16]. The robot's control loop is a cycle that alternates sensing and acting phases in discrete time steps. Formally, the task environment is defined as follows:

**Environment**: it consists of a square arena ($1m$ x $1m$) with a light source positioned in one corner.

**Performance measure**: in the beginning of the experiment, the robot is located in a random position close to the opposite corner of the arena with respect to the light. Then, given a certain execution time $T$, the robot must satisfy two properties to successfully achieve the task: it must go towards the light, until a clap is perceived; after that, it must move away from the light.[4] We define the performance as a function of an error $E \in [0, 1]$, to be minimised. The smaller is the error, the better is the robot performance. The error function is given by a weighted mean including the phototaxis contribute and the antiphototaxis one: at each time step $t \in \{1, \ldots, T\}$, the robot is rewarded if it is moving in the correct direction with respect to the light. Let $t_c$ be the time instant at which the clap is performed. We can write the error function $E$ as follows:

$$E = \alpha \Big(1 - \frac{\sum_{i=1}^{t_c} s_i}{t_c}\Big) + \Big(1 - \alpha\Big)\Big(1 - \frac{\sum_{i=t_c+1}^{T} s_i}{T - t_c}\Big),$$

where:

$$\forall i \in \{1, \ldots, t_c\},\ s_i = \begin{cases} 1 & \text{if the robot goes towards to the light at step } i \\ 0 & \text{otherwise} \end{cases}$$

$$\forall i \in \{t_c+1, \ldots, T\},\ s_i = \begin{cases} 1 & \text{if the robot moves away from the light at step } i \\ 0 & \text{otherwise} \end{cases}$$

---
[4]We have also recently developed a robot behaviour in which phototaxis and antiphototaxis

**Sensors**: light sensors enable the robot to assess its position with respect to the light. The robot has a circular body and it is equipped with eight light sensors whose values are combined in such a way that the robot can perceive the light in eight different sectors with equal angle of $\frac{\pi}{4}$. We denote the eight possible sensor readings by assigning each of them a numerical identifier, from 1 (North) to 8 (North-West), clockwise. Moreover, the robot is equipped with a sound sensor whose value is 1, if the clap is perceived in the current time step, or 0 otherwise.

In this experiment, wheel speeds are either set to zero or to a predefined, constant value, to make it possible to associate a binary value to their control. In this way, it is possible for the robot to move in any direction by simply setting the proper combination of two binary values.

### 4.1 BN setup

The BN implementing the robot program is subject to a synchronous and deterministic update; moreover, it is also synchronous with the robot. At each time step, three operations are performed in order:

1. The sensor readings are encoded into input values.
2. The network's state is updated.
3. The value of the output nodes is read, encoded and used to operate on the actuators.

Network size is a design parameter which could be set manually or defined by an automatic procedure. In this work, we set the number of network nodes to 20, as this size provides a trade-off between computation cost in simulating the network and size of the network state space.

In order to map sensors and actuators onto BN's inputs and outputs, we use one input node per sensor. The value of node $x_1$ of the BN is set depending on are alternated if the clap is emitted more than once. This robot is still under testing.

the binary value read on the sound sensor. Four input nodes ($x_2$, $x_3$, $x_4$, $x_5$) are connected to light sensors whose eight sensor readings are coded by means of a Gray code of four Boolean variables.[5] The values of nodes $x_6$ and $x_7$ (output nodes) are used to control the wheel actuators.

## 4.2 Design of BN robot

The local search we implemented is a simple stochastic descent, in which moves can change one value in a node function's truth table. While, in principle, it would be possible to use more elaborated search strategies, as well as moves involving also the BN topology, this choice is mainly motivated by the goal of checking whether a simple 'evolutionary walk' (single flip fitter mutant) across the Boolean functions space can produce a BN able to make the robot attaining the desired behaviour. Moreover, this choice makes it possible to study in detail the error landscape and devise further improved search techniques. The initial connections among nodes are randomly generated with $K = 3$ (no self-connections) and are kept fixed during the search. The initial Boolean functions are generated by setting the 0/1 values in the truth tables uniformly at random. A point in the search space is thus a 20-nodes BN with given topology; a move consists of one flip in one function's truth table. The search strategy is that of stochastic descent: a neighbour is chosen at random (random entry in the truth table of a randomly chosen node) and accepted if the corresponding BN has an evaluation not worse than the current one. We executed 30 independent experiments, each corresponding to a different initial BN. In each experiment we trained the robot by means of local search in a simulated environment. The set of initial conditions form the training set (the same for each experiment), composed of 30 different positions of the robot. At the end of the training process, we tested the BNs obtained in a simulated environment on initial conditions different from those of the training set. Considering the size of the arena and the speed of robot wheels, we empirically estimated that 1000 time steps are enough to let the robot achieve the task. Each item in the training set is characterised by an initial position and orientation of the robot with respect to the light, chosen at random. In addition, also the instant the clap is triggered is chosen at random in $\{500, \ldots, 650\}$. In order to train the BN-robot so as to make it robust with respect to both sensor and actuator noise, at a random instant we impose the robot rotation of an angle $\theta$ randomly chosen in $[-\pi, \pi]$. This external change has the effect of forcing the robot to correct its direction. The local search was run for 25000 iterations. As from preliminary experiments we made and common experience in evolutionary robotics [17], we split the design goal into two, subsequent, sub-goals: in the first 5000 iterations of the optimisation algorithm, the simulations last only 500 time steps without any clap. The goal is to obtain agent programs able to perform phototaxis and to gain robustness against noise; in the subsequent 20000 iterations, the simulations last 1000 time steps and they involve the clap. Thus, the idea is to train the agent incrementally. The best BN resulting from this process was then ported into a real robotic platform for assessing its performance.

Results of training and testing are reported in the boxplots of Figures 4(a) and 4(b), respectively. Boxplots graphically summarise statistics of the results achieved in each experiment. The bold line of the rectangle denotes the median value, while its uppermost and lowermost sides denote the 1st and 3rd quartile, respectively. We can observe that 4 out of 30 runs were successful in the training phase and they attained a good performance also in testing, three of which with a median error less than 0.11. In the other runs, the BN-robot was able to perform correctly only phototaxis, hence a performance with an error around 0.5. It is important to remark that while the design goal is quite hard, somehow surprisingly the stochastic descent was able to attain a success ratio of 13%. The best performing BN has also been tested in a real setting with an e-puck robot and tests confirmed that the robot is able to achieve the task from different initial conditions and clap instants. A further remarkable property we observed is that the robot is also able to react to external alterations of its position and correct its trajectory. The analysis of the features of the best performing BN is omitted for lack of space; further details on the BN used for testing the robot, along with a movie of a typical run of the e-puck BN-robot can be found as online supplementary material [22].

---
[5]The Gray code is a binary numeral system in which two successive code words differ in only one bit.

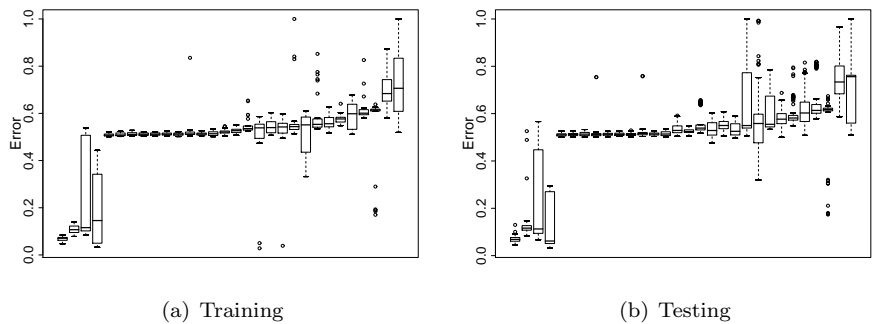

(a) Training  (b) Testing

Figure 4: Results of the BN-robot testing. BNs are ordered by median error in training.

## 5 Conclusion and future work

In this paper we have introduced Boolean network robotics, in which robots and artificial agents in general are programmed by means of BN models. We have shown that a BN can be designed through stochastic local search so as to create a robot able to alternate phototaxis and antiphototaxis behaviour depending on a sound signal.

In future work, we plan to study different kinds of input/output mappings and to exploit the complex dynamics of BNs in more challenging robot tasks, for example in changing environments. In addition, different update schemes for BNs can be explored, as well as other GRN models. Besides this, a principled theoretical and experimental framework to study the dynamics of BNs interacting with the environment—i.e., an entity with its own dynamics—is subject of ongoing work.

# Acknowledgements

Mattia Manfroni acknowledges support from "Seconda Facoltà di Ingegneria", *Alma Mater Studiorum* Università di Bologna. Carlo Pinciroli acknowledges support from ASCENS, a project funded by the Future and Emerging Technologies programme of the European Commission. Mauro Birattari acknowledges support from the fund for scientific research F.R.S. – FNRS of the French Community of Belgium.